\documentclass[final]{cvpr}

\usepackage{times}
\usepackage{epsfig}
\usepackage{graphicx}
\usepackage{amsmath}
\usepackage{amssymb}
\usepackage{marvosym}
\usepackage{booktabs}

\usepackage[ruled,vlined]{algorithm2e}
\usepackage[pagebackref=true,breaklinks=true,colorlinks,bookmarks=false]{hyperref}
\newcommand{\dtoprule}{\specialrule{1pt}{0pt}{0.4pt}%
            \specialrule{0.3pt}{0pt}{\belowrulesep}%
            }

\DeclareMathOperator*{\argmax}{arg\,max}
\def\cvprPaperID{6} 
\def\confYear{CVPR 2021}

\begin{document}

\title{DAMSL: Domain Agnostic Meta Score-based Learning}

\author{John Cai \\
Princeton University \\
  {\tt jjcai@alumni.princeton.edu}
\and
Bill Cai \\
Massachusetts Institute of Technology \\
  {\tt billcai@alum.mit.edu} 
\and 
Shen Sheng Mei \\
Pensees Pte Ltd \\
  {\tt jane.shen@pensees.ai}
}

\maketitle

\begin{abstract}
  In this paper, we propose Domain Agnostic Meta Score-based Learning (DAMSL), a novel, versatile and highly effective solution that delivers significant out-performance over state-of-the-art methods for cross-domain few-shot learning. We identify key problems in previous meta-learning methods over-fitting to the source domain, and previous transfer-learning methods under-utilizing the structure of the support set.  The core idea behind our method is that instead of directly using the scores from a fine-tuned feature encoder, we use these scores to create input coordinates for a domain agnostic metric space. A graph neural network is applied to learn an embedding and relation function over these coordinates to process all information contained in the score distribution of the support set. We test our model on both established CD-FSL benchmarks and new domains and show that our method overcomes the limitations of previous meta-learning and transfer-learning methods to deliver substantial improvements in accuracy across both smaller and larger domain shifts.
\end{abstract}

\section{Introduction}

Few-shot learning methods promise to solve one of the most challenging issues in deep learning: the reliance on copious amounts of labelled examples to achieve high accuracies. By doing so, we can achieve cost savings and accurately classify rare classes of images (e.g. plane crashes) where labelled examples are limited. The problem, however, is that few-shot learning methods fail to perform well when there is a domain-shift. Hence, the practical applications of few-shot learning are severely limited as the few-shot models trained on well-labelled and well-structured research datasets cannot be applied to domains in industry.

The above problem is exacerbated under sharp domain shifts, as shown in the Broader Study of Cross-Domain Few-Shot Learning (BSCD-FSL) \cite{guo2020broader}. The study found that many few-shot learning methods significantly under-performed compared to transfer-learning as few-shot learning overfitted to the source domain. While transfer-learning methods did perform better, they omitted distributional information contained in the each episode's support set is omitted. This is clearly sub-optimal given the need to maximally use information in the sparse setting \cite{zhang2019variational} \cite{triantafillou2017few}.

To solve the above issues, we propose Domain Agnostic Meta Score-based Learning (DAMSL). The fundamental idea behind our method is to apply transfer-learning to prevent over-fitting to the source domain, while using metric-learning to exploit the information in each episode's support samples. Furthermore, as metric-learners built on image features are shown to suffer greatly from overfitting to the source domain, we make our metric-learner domain-agnostic by fitting to pre-softmax classification scores from fine-tuned feature encoders. 

In our work, we use the BSCD-FSL benchmark \cite{guo2020broader} and augment it with 4 more test domains for further comparisons. We demonstrate the superiority of our method over existing methods across these 8 distinct test domains, and show a new research direction for score-based boosting in the few-shot classification setting. 

\begin{figure*}[t]
\begin{center}
   \includegraphics[width=0.7\linewidth]{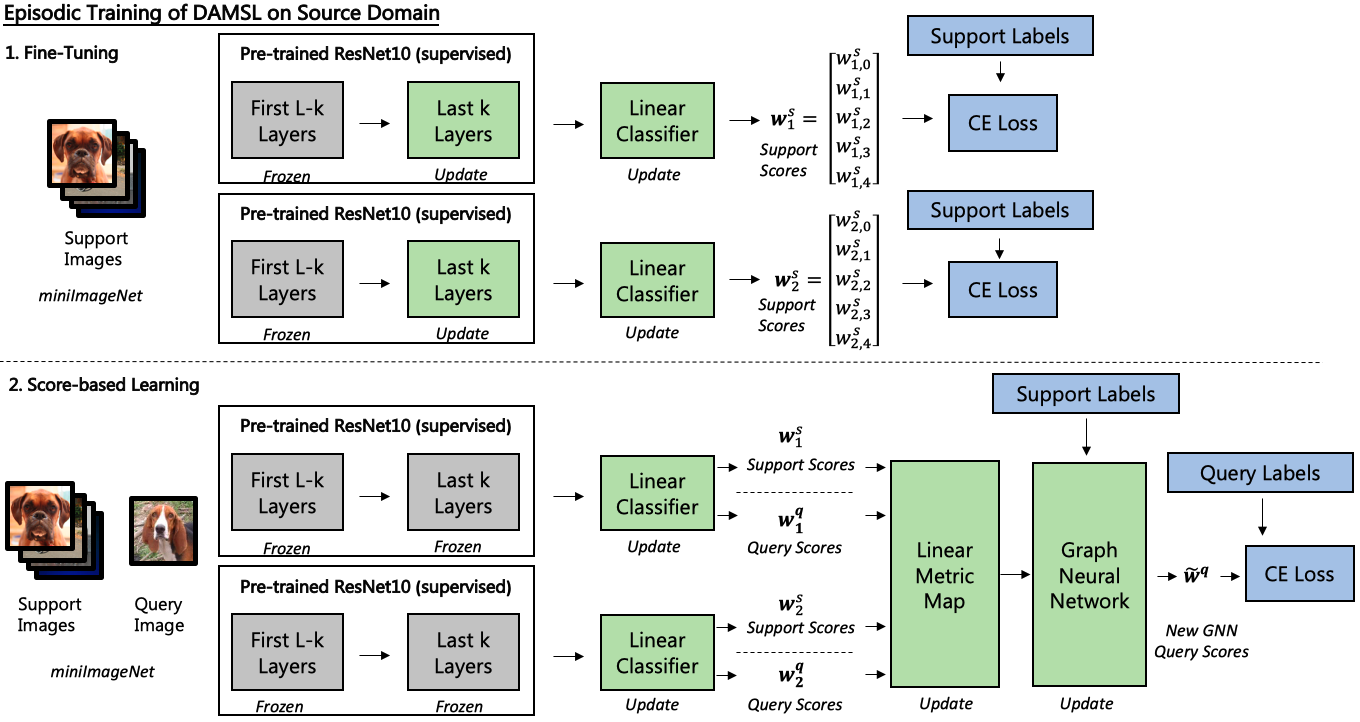}
\end{center}
   \caption{Episodic training on miniImagenet (source domain) for our Proposed DAMSL Model.}
\label{fig:DAMSL}
\end{figure*}

\section{Relevant Work}

Metric-based methods, such as prototypical networks \cite{snell2017prototypical}, aim to learn a metric function \(\phi_m\) that can be used to classify query images based on their relations to the images in the support set. The key metric-based method that we use is \textbf{Graph Neural Network} (GNN) as graph-based convolutions can create more flexible representations \cite{bronstein2017geometric}.

Transfer learning involves reusing features learned from base classes \cite{pan2009survey}, typically by fine-tuning a pre-trained model. A simple extension of fine-tuning would be to \textbf{learn to fine-tune}. Methods such as MAML \cite{finn2017model} learn an internal representation that can be fine-tuned in a few gradient steps.

\section{Methodology}

The training process during each episode for our model is shown in Figure \ref{fig:DAMSL}. On test domains, the same fine-tuning process occurs over the labelled support set, but with gradient updates only within episodes and not between episodes.

\subsection{Score-based Metric Learning}

Given only a sparse support set from the test domain, it is difficult to precisely fit the feature encoder in a way that neither overfits nor underfits \cite{nakamura2019revisiting}. Creating a hold-out validation set is also prohibitively costly under such conditions.  

We begin by fine-tuning a feature vector to obtain \(\Tilde{\phi}_f (X_s) \in \mathbb{R}^{512}\). Then, we take the linear classifier \(\Tilde{\phi}_c \) to produce a pre-softmax score vector \(\Tilde{\phi}_c (\Tilde{\phi}_f (X_i)) \in \mathbb{R}^5\), which corresponds to our 5-way classification problem. A typical transfer-learning approach would directly use the score vector for prediction \(\hat{Y}_q = \argmax  (\Tilde{\phi}_c (\Tilde{\phi}_f (X_q))  \)

Instead, we post-process the score vector by using a metric-learning network \(\phi_m\). Formally, this gives us: \(\hat{Y}_q = \argmax (\phi_m (Y_s, \Tilde{\phi}_c (\Tilde{\phi}_f (X_q)), \Tilde{\phi}_c (\Tilde{\phi}_f (X_s))))\) Thus, we explicitly incorporate information contained in the predictions we can make on the support sets \(X_s\) and how these predictions correspond to the labels \(Y_s\). Any biases found in the feature encoder \(\Tilde{\phi}_f \) and linear classifier \(\Tilde{\phi}_c\)  can be corrected by inferring a distribution from the support set scores, and using that distribution to match the scores of our query samples. This reduces the reliance on the initial fine-tuning process as our decision boundaries from the initial feature encoder are replaced by a metric-based decision boundary constructed from the proximity of the query sample to the support classes. Moreover, the scores form a domain agnostic basis for metric learning because the way that scores relate should not differ significantly across domains.

\subsection{Graph Neural Network}
The meta-learning module we use is the Graph Neural Network (GNN). We follow the formulation of the GNN for the few-shot problem in \cite{satorras2018few}. In brief, a GNN acts on local operators of a graph \(G = (V,E)\), which for the few-shot learning case is fully connected. A graph convolution layer \(GC(.)\) \cite{satorras2018few} is performed with linear operations on local signals. Formally, we have:
\begin{equation}
    GC(S^{k}) = f \Bigg (\sum_{B\in \mathcal{A}}B S^k \theta^{k}_{B,q} \Bigg ) , \;  q = d_1, ...,d_{k+1}  
    \label{eq:graphconv}
\end{equation}

In the few-shot learning formulation, we can learn the edfe features using the current hidden vertex \cite{satorras2018few}. We apply a Multi-layer Perceptron (MLP) that takes in the absolute difference between the the output vectors of vertices in the graph \cite{kearnes2016molecular} \cite{gilmer2017neural}. Formally, we have:
\begin{equation}
    \Tilde{A}_{i,j}^{k} = \gamma(S_i^k, S_j^k) = MLP(|S_i^k - S_j^k |)
    \label{eq:edge}
\end{equation}
These learned edge features are used to propagate information in the graph through the graph convolution in equation \ref{eq:graphconv}. Initial vertex features are constructed by taking the score projections and one-hot encoding of labels for the support set or a uniform distribution for the query samples.

\begin{table*}[t]
\makebox[\textwidth][c]{
\resizebox{\textwidth}{!}{%
    \begin{tabular}{ccccccc}
\dtoprule
 \textbf{Methods}   & \multicolumn{3}{c}{\textbf{EuroSAT}} & \multicolumn{3}{c}{\textbf{CropDisease}} \\
\cmidrule(lr){2-4}\cmidrule(lr){5-7}
   & 5-way 5-shot  & 5-way 20-shot & 5-way 50-shot  & 5-way 5-shot  & 5-way 20-shot & 5-way 50-shot \\ \midrule
ProtoNet\textsuperscript{\Cross} & 73.29\% \(\pm\) 0.71\% & 82.27\% \(\pm\) 0.57\% & 80.48\% \(\pm\) 0.57\% & 79.72\% \(\pm\) 0.67\% & 88.15\% \(\pm\) 0.51\% & 90.81\% \(\pm\) 0.43\% \\
TransFT\textsuperscript{\Cross} & 81.76\% \(\pm\) 0.48\% & 87.97\% \(\pm\) 0.42\% & 92.00\% \(\pm\) 0.56\% &  90.64\% \(\pm\) 0.54\% & 95.91\% \(\pm\) 0.72\% & 97.48\% \(\pm\) 0.56\% \\
L-Ensem v1 & 74.64\% \(\pm\) 0.67\% & 85.52\% \(\pm\) 0.53\% & 90.38\% \(\pm\) 0.35\% & 84.65\% \(\pm\) 0.60\% & 94.40\% \(\pm\) 0.36\% & 96.89\% \(\pm\) 0.24\%  \\
DAMSL v1 & 85.93\% \(\pm\) 0.68\% & 95.18\% \(\pm\) 0.35\% & 97.73\% \(\pm\) 0.25\% & 95.03\% \(\pm\) 0.42\% & 99.19\% \(\pm\) 0.14\% & 99.75\% \(\pm\) 0.08\%  \\
L-Ensem v2 & 81.02\% \(\pm\) 0.62\% & 90.01\% \(\pm\) 0.37\% & 93.28\% \(\pm\) 0.30\% & 90.68\% \(\pm\) 0.51\% & 97.20\% \(\pm\) 0.26\% & 98.89\% \(\pm\) 0.16\%  \\
DAMSL v2 & \textbf{90.84\% \(\pm\) 0.54\%} & \textbf{96.13\% \(\pm\) 0.29\%} & \textbf{98.15\% \(\pm\) 0.18\%} & \textbf{97.30\% \(\pm\) 0.32\%} &\textbf{ 99.36\% \(\pm\) 0.14\%} & \textbf{99.73\% \(\pm\) 0.09\%}  \\
\toprule
TransFT (Aug) & 82.80\% \(\pm\) 0.60\% & 90.38\% \(\pm\) 0.67\% & 93.48\% \(\pm\) 0.57\% & 92.51\% \(\pm\) 0.84\% & 97.33\% \(\pm\) 0.43\% & 98.40\% \(\pm\) 0.41\%  \\
L-Ensem v1 (Aug) & 77.98\% \(\pm\) 0.66\% & 89.43\% \(\pm\) 0.39\% & 93.56\% \(\pm\) 0.31\% & 88.84\% \(\pm\) 0.54\% & 97.11\% \(\pm\) 0.23\% & 98.83\% \(\pm\) 0.18\%  \\
FT-GNN v1 (Aug) & 82.29\% \(\pm\) 0.63\% & 92.73\% \(\pm\) 0.63\% & 93.78\% \(\pm\) 0.63\% & 94.09\% \(\pm\) 0.46\% & 98.31\% \(\pm\) 0.35\% & 98.95\% \(\pm\) 0.25\%  \\
S-Proto v1 (Aug) & 80.32\% \(\pm\) 0.58\% & 89.52\% \(\pm\) 0.40\% & 93.39\% \(\pm\) 0.27\% & 91.43\% \(\pm\) 0.47\% & 97.53\% \(\pm\) 0.37\% & 98.79\% \(\pm\) 0.26\%  \\
DAMSL v1 (Aug) & 87.30\% \(\pm\) 0.68\% & 96.53\% \(\pm\) 0.28\% & 98.37\% \(\pm\) 0.18\% & 96.01\% \(\pm\) 0.40\% & \textbf{99.61\% \(\pm\) 0.09\%} & 99.85\% \(\pm\) 0.06\%  \\
L-Ensem v2 (Aug) & 83.68\% \(\pm\) 0.54\% & 91.61\% \(\pm\) 0.34\% & 94.75\% \(\pm\) 0.25\% & 92.66\% \(\pm\) 0.45\% & 98.08\% \(\pm\) 0.20\% & 99.14\% \(\pm\) 0.11\%  \\
DAMSL v2 (Aug) & \textbf{91.59\% \(\pm\) 0.49\%} & \textbf{96.99\% \(\pm\) 0.24}\% & \textbf{98.60\% \(\pm\) 0.15\%} & \textbf{97.43\% \(\pm\) 0.31\%} & \textbf{99.61\% \(\pm\) 0.10\%} & \textbf{99.87\% \(\pm\) 0.05\%}  \\
\bottomrule
\end{tabular}}%
}
\\[1ex]
\makebox[\textwidth][c]{
\resizebox{\textwidth}{!}{%
    \begin{tabular}{ccccccc}
\dtoprule
 \textbf{Methods}   & \multicolumn{3}{c}{\textbf{ChestX}} & \multicolumn{3}{c}{\textbf{ISIC}} \\
\cmidrule(lr){2-4}\cmidrule(lr){5-7}
   & 5-way 5-shot  & 5-way 20-shot & 5-way 50-shot  & 5-way 5-shot  & 5-way 20-shot & 5-way 50-shot \\ \midrule
ProtoNet\textsuperscript{\Cross} & 24.05\% \(\pm\) 1.01\% & 28.21\% \(\pm\) 1.15\% & 29.32\% \(\pm\) 1.12\% & 39.57\% \(\pm\) 0.57\% & 49.50\% \(\pm\) 0.55\% & 51.99\% \(\pm\) 0.52\% \\
TransFT\textsuperscript{\Cross} & 26.09\% \(\pm\) 0.96\% & 31.01\% \(\pm\) 0.59\% & 36.79\% \(\pm\) 0.53\% & 49.68\% \(\pm\) 0.36\% & 61.09\% \(\pm\) 0.44\% & 67.20\% \(\pm\) 0.59\%  \\
L-Ensem v1 & 25.20\% \(\pm\) 0.43\% & 30.62\% \(\pm\) 0.45\% & 35.82\% \(\pm\) 0.47\% & 46.55\%  \(\pm\) 0.61\% &  59.14\% \(\pm\) 0.61\% & 65.35\% \(\pm\) 0.59\% \\
DAMSL v1 & 25.99\% \(\pm\) 0.50\% & 33.47\% \(\pm\) 0.54\% & 38.37\% \(\pm\) 0.56\% & 50.68\%  \(\pm\) 0.76\% & 68.58\% \(\pm\) 0.70\% & 75.55\% \(\pm\) 0.58\%  \\
L-Ensem v2 & 26.38\% \(\pm\) 0.45\% & 33.46\% \(\pm\) 0.51\% & 39.81\% \(\pm\) 0.53\% & 51.93\%  \(\pm\) 0.62\% &  64.21\% \(\pm\) 0.60\% & 70.28\% \(\pm\) 0.57\% \\
DAMSL v2 & \textbf{27.22\% \(\pm\) 0.49\%} & \textbf{35.41\% \(\pm\) 0.56\%} & \textbf{42.74\% \(\pm\) 0.62\%} & \textbf{57.35\%  \(\pm\) 0.78\%} & \textbf{70.32\% \(\pm\) 0.70\%} & \textbf{77.40\% \(\pm\) 0.65\%}  \\ 
\toprule
TransFT (Aug) & \textbf{29.23\% \(\pm\) 0.46\%} & 36.25\% \(\pm\) 0.55\% & 40.69\% \(\pm\) 0.56\% & 51.54\% \(\pm\) 0.64\% & 62.72\% \(\pm\) 0.62\% & 69.68\% \(\pm\) 0.59\% 
\\
L-Ensem v1 (Aug) & 26.84\% \(\pm\) 0.44\% & 34.62\% \(\pm\) 0.48\% & 40.23\% \(\pm\) 0.56\% & 48.97\%  \(\pm\) 0.65\% & 62.99\% \(\pm\) 0.60\% & 70.32\% \(\pm\) 0.57\% \\
FT-GNN v1 (Aug) & 26.79\% \(\pm\) 0.50\% & 35.39\% \(\pm\) 0.60\% & 35.34\% \(\pm\) 0.54\% & 52.13\% \(\pm\) 0.84\% & 65.37\% \(\pm\) 0.73\% & 62.68\% \(\pm\) 0.65\%  \\
S-Proto v1 (Aug) & 27.55\% \(\pm\) 0.44\% & 35.37\% \(\pm\) 0.57\% & 41.56\% \(\pm\) 0.56\% & 50.98\% \(\pm\) 0.65\% & 63.58\% \(\pm\) 0.61\% & 71.47\% \(\pm\) 0.54\%  \\
DAMSL v1 (Aug) & 28.08\% \(\pm\) 0.50\% & \textbf{37.70\% \(\pm\) 0.57\%} & \textbf{43.04\% \(\pm\) 0.66\%} & 53.50\% \(\pm\) 0.79\% & 70.31\% \(\pm\) 0.72\% & 78.41\% \(\pm\) 0.66\%  \\
L-Ensem v2 (Aug) & 28.26\% \(\pm\) 0.46\% & 35.91\% \(\pm\) 0.52\% & 41.00\% \(\pm\) 0.57\% & 52.76\%  \(\pm\) 0.62\% & 64.99\% \(\pm\) 0.59\% & 71.92\% \(\pm\) 0.55\% \\
DAMSL v2 (Aug) & 28.86\% \(\pm\) 0.52\% & 37.04\% \(\pm\) 0.61\% & 42.87\% \(\pm\) 0.65\% & \textbf{57.15\% \(\pm\) 0.76\%} & \textbf{70.87\% \(\pm\) 0.72\%} & \textbf{78.98\% \(\pm\) 0.62\%}  \\
\bottomrule
\end{tabular}}%
}
\begin{footnotesize}
\Cross  -  as reported in \cite{guo2020broader}. Aug - with data augmentation during fine-tuning. \textbf{Bold} - Best performing in category. 
\end{footnotesize} 
\vskip 1mm
\caption{Results on BSCD-FSL Benchmark. Includes ablation studies and results from prior work. }
\label{tab:singlesource}%
\end{table*}%

\subsection{Backbone Variants}
We experiment with two variants of the DAMSL model with different feature backbones. In \textbf{DAMSL v1}, we have a ResNet10 pre-trained using supervised learning and first-order MAML \cite{nichol2018first} In \textbf{DAMSL v2}, we have two ResNet10 pre-trained using supervised learning but with different optimization strategies - one trained using Adam while other trained using SGD with momentum. 

\section{Results and Discussion}

\subsection{Experimental Setup}
First, we test our model on the BSCD-FSL benchmark. We train on miniImagenet and test on CropDisease \cite{mohanty2016using}, EuroSAT \cite{helber2019eurosat}, ISIC \cite{tschandl2018ham10000} \cite{codella2019skin} and ChestX \cite{wang2017chestx} (in order of decreasing similarity). CropDisease covers plant disease, EuroSAT covers satellite images, ISIC covers dermoscopic skin lesion images and ChestX covers chest X-ray images. 

4 other new datasets are also included: Places \cite{zhou2017places}, Describable Textures Dataset (DTD) \cite{cimpoi2014describing}, CIFAR-100 \cite{krizhevsky2009learning} and Caltech256 \cite{griffin2007caltech}. While the images in these other datasets are all natural images, they contain very different types of classification tasks from miniImagenet. For instance, DTD requires the model to recognize textures in many contexts \cite{cimpoi2014describing}. The Github repo can be found here: \href{https://github.com/johncai117/DAMSL}{Github link}.

\subsection{Results on the BS-CDFSL Benchmark}

From Table \ref{tab:singlesource}, DAMSL with both feature backbones outperform previous methods, even without using data augmentation. We see that DAMSL v1 achieves an average accuracy of \textbf{72.13\%} while DAMSL v2 achieves an average accuracy of \textbf{74.34\%}. With data augmentation, DAMSL v1 (Aug) achieves an average accuracy of \textbf{74.06\%} while DAMSL v2 (Aug) achieves an average accuracy of \textbf{74.99\%}.

\begin{table*}[t]
\makebox[\textwidth][c]{
\resizebox{\textwidth}{!}{%
    \begin{tabular}{ccccccc}
\dtoprule
 \textbf{Methods}   & \multicolumn{3}{c}{\textbf{DTD}} & \multicolumn{3}{c}{\textbf{CIFAR-100}} \\
\cmidrule(lr){2-4}\cmidrule(lr){5-7}
   & 5-way 5-shot  & 5-way 20-shot & 5-way 50-shot  & 5-way 5-shot  & 5-way 20-shot & 5-way 50-shot \\ \midrule
TransFT (Aug) & 62.17\% \(\pm\) 0.74\% & 73.49\% \(\pm\) 0.61\% & 79.25\% \(\pm\) 0.55\% & 65.89\% \(\pm\) 0.77\% & 77.60\% \(\pm\) 0.64\% & 83.64\% \(\pm\) 0.54\%  \\
L-Ensem v1 (Aug) & 55.80\% \(\pm\) 0.73\% & 69.20\% \(\pm\) 0.69\% & 76.39\% \(\pm\) 0.59\% & 62.07\% \(\pm\) 0.77\% & 77.85\% \(\pm\) 0.61\% & 84.17\% \(\pm\) 0.51\%  \\
DAMSL v1 (Aug) & 58.29\% \(\pm\) 0.89\% & 77.72\% \(\pm\) 0.71\% & 85.44\% \(\pm\) 0.58\% & 67.64\% \(\pm\) 0.93\% & 86.68\% \(\pm\) 0.64\% & 93.06\% \(\pm\) 0.42\%  \\
L-Ensem v2 (Aug) & 60.01\% \(\pm\) 0.74\% & 73.73\% \(\pm\) 0.65\% & 79.99\% \(\pm\) 0.55\% & 66.01\% \(\pm\) 0.42\% & 80.67\% \(\pm\) 0.59\% & 86.23\% \(\pm\) 0.46\%  \\
DAMSL v2 (Aug) & \textbf{68.39\% \(\pm\) 0.89\%} & \textbf{81.64\% \(\pm\) 0.68\%} & \textbf{87.14\% \(\pm\) 0.68\%} & \textbf{76.56\% \(\pm\) 0.85\%} & \textbf{88.47\% \(\pm\) 0.57\%} & \textbf{93.92\% \(\pm\) 0.39\%}  \\
\bottomrule
\end{tabular}}%
}
\\[1ex]
\makebox[\textwidth][c]{
\resizebox{\textwidth}{!}{%
    \begin{tabular}{ccccccc}
\dtoprule
 \textbf{Methods}   & \multicolumn{3}{c}{\textbf{Places}} & \multicolumn{3}{c}{\textbf{Caltech256}} \\
\cmidrule(lr){2-4}\cmidrule(lr){5-7}
   & 5-way 5-shot  & 5-way 20-shot & 5-way 50-shot  & 5-way 5-shot  & 5-way 20-shot & 5-way 50-shot \\ \midrule
TransFT (Aug) & 67.50\% \(\pm\) 0.75\% & 76.17\% \(\pm\) 0.67\% & 80.98\% \(\pm\) 0.57\% & 75.32\% \(\pm\) 0.70\% & 84.15\% \(\pm\) 0.58\% & 88.37\% \(\pm\) 0.47\%  \\
L-Ensem v1 (Aug) & 70.78\% \(\pm\) 0.92\% & 74.45\% \(\pm\) 0.65\% & 79.85\% \(\pm\) 0.57\% & 70.32\% \(\pm\) 0.72\% & 83.20\% \(\pm\) 0.55\% & 87.59\% \(\pm\) 0.48\%  \\
DAMSL v1 (Aug) & 71.34\% \(\pm\) 0.85\% & 84.30\% \(\pm\) 0.67\% & 89.50\% \(\pm\) 0.55\% & 76.63\% \(\pm\) 0.87\% & 91.44\% \(\pm\) 0.49\% & 95.06\% \(\pm\) 0.37\%  \\
L-Ensem v2 (Aug) & \textbf{75.45\% \(\pm\) 0.80\%} & 75.87\% \(\pm\) 0.65\% & 82.13\% \(\pm\) 0.53\% & 76.31\% \(\pm\) 0.68\% & 87.59\% \(\pm\) 0.44\% & 90.93\% \(\pm\) 0.35\%  \\
DAMSL v2 (Aug) & 75.42\% \(\pm\) 0.80\% & \textbf{84.74\% \(\pm\) 0.60\%} & \textbf{90.43\% \(\pm\) 0.51\%} & \textbf{87.44\% \(\pm\) 0.70\%} & \textbf{94.53\% \(\pm\) 0.36\%} & \textbf{96.89\% \(\pm\) 0.24\%}  \\
\bottomrule
\end{tabular}}%
}
\begin{footnotesize}
 Aug - with data augmentation during fine-tuning. \textbf{Bold} - Best performing in category. 
\end{footnotesize} 
\vskip 1mm
\caption{Results on Additional Test Domains}
\label{tab:other}%
\end{table*}%

We focus on DAMSL v2 in our comparison with other methods as it has the highest performance. DAMSL v2 (Aug) outperforms TransFT by 6.86\% and ProtoNet by 15.21\%. From Table \ref{tab:singlesource}, we see that DAMSL (Aug) v2 still significantly outperforms TransFT (Aug), with the exception of 5-way 5-shot Chest-X. In terms of average accuracy, DAMSL v2 (Aug) outperforms TransFT (Aug) by 3.84\%.

Our method is competitive with typical supervised learning on domains closer to the source domain. Typical supervised learning models for EuroSAT and CropDisease have achieved 98.57\% and 99.35\% respectively \cite{helber2019eurosat} \cite{mohanty2016using}. At our 50-shot results for DAMSL v2 (Aug), we achieve 98.60\% and 99.87\% respectively on EuroSAT and CropDisease. 

\subsection{Ablation Studies on BSCD-FSL}
To investigate the effect of each component of separately, we include a Linear Ensemble (L-Ensem), a Fine-Tuned encoder + GNN (FT-GNN), and Score-based ProtoNets (S-Proto) in Table \ref{tab:singlesource}. The L-Ensem is a simple addition of the post-softmax scores from the two fine-tuned feature encoders, to see the performance from a simple fine-tuned ensemble. FT-GNN is directly fitted to a feature vector that has been fine-tuned on the support set, to demonstrate the performance boost from score-based learning. The Score-based ProtoNets replaces the GNN module with an embedding MLP and a nearest centroid classifier, to demonstrate the additional gains from a GNN module.

Furthermore, we see that the score-based metric delivers an improvement when used in conjunction with a simple ProtoNets. On all tasks, the S-Proto delivers a better performance compared to L-Ensem. However, it still does not match up to the performance of our proposed DASML model. We attribute this to the GNN's more flexible representations, and the fact that it exploits the full distribution of scores rather than just the mean value of scores. We also demonstrate the value of score-based metric learning as FT-GNN performs substantially worse than DAMSL. This is in line with the expectation that the GNN is trained to interpret domain-specific features, and thus fails on distant domains.

Looking at average accuracy using v1, we observe that L-Ensem yields \textbf{69.23\%}, FT-GNN yields \textbf{69.82\%}, S-Proto yields \textbf{70.12\%}, DAMSL v1 yields \textbf{74.06\%}. This shows that both parts of DAMSL are most useful when jointly applied.  

\subsection{Results on other Test Domains}

From Table \ref{tab:other}, we clearly see that DAMSL v2 delivers out-performance over the previous baselines across almost all settings and all shots, with the only exception of 5-shot setting for Places. In terms of average accuracy, DAMSL v1 achieves \textbf{81.48\%} while DAMSL v2 achieves \textbf{85.46\%} . These values are considerably higher \((>5\%\)) than the linear ensembles, which yields \textbf{74.31\%} and \textbf{77.91\%} respectively. The results clearly validate our method as DAMSL performs better than previous methods on data-sets other than those in the BSCD-FSL benchmark.

\begin{figure}[t]
\begin{center}
   \includegraphics[width=0.95\linewidth]{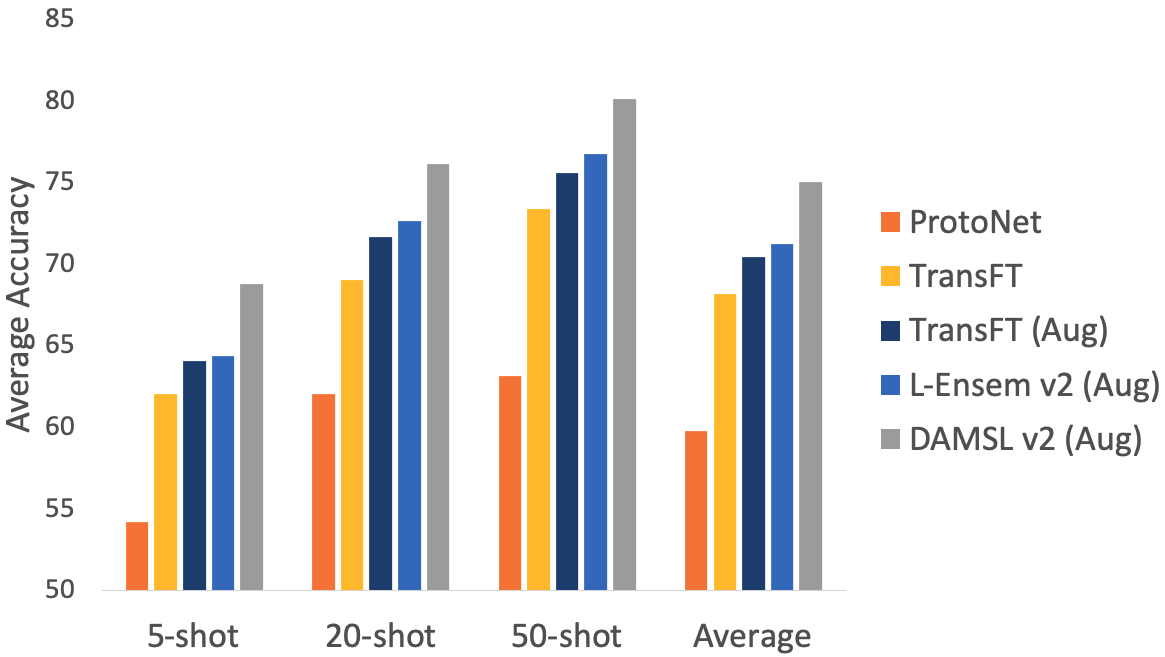}
\end{center}
   \caption{Summary of performance on BSCD-FSL}
\label{fig:Summary}
\end{figure}

\section{Conclusion}

We propose Domain Agnostic Meta Score-based Learning (DAMSL) to address the Cross-Domain Few-Shot Learning problem. On BSCD-FSL, DAMSL v2 achieves \textbf{74.99\%} accuracy, which significantly outperforms previous best-performing meta-learning and transfer-learning methods by \textbf{15.21\%}  and \textbf{6.86\%} respectively. From Figure \ref{fig:Summary}, we also see that the performance gains from our method are substantially greater than gains from including data augmentation or adding another feature encoder. Moreover, on the 4 other test domains beyond BSCD-FSL, our method continues to consistently outperform strong baselines.

Ultimately, we not only decisively address the CD-FSL problem, but we also outline a new strand of classification boosting modules that can be attached to any existing model to self-correct initial classification scores by utilizing the distributional information of scores from labelled samples.

{\small
\bibliographystyle{ieee_fullname}
\bibliography{egbib}

\begin{thebibliography}{10}\itemsep=-1pt

\bibitem{bronstein2017geometric}
Michael~M Bronstein, Joan Bruna, Yann LeCun, Arthur Szlam, and Pierre
  Vandergheynst.
\newblock Geometric deep learning: going beyond euclidean data.
\newblock {\em IEEE Signal Processing Magazine}, 34(4):18--42, 2017.

\bibitem{cimpoi2014describing}
Mircea Cimpoi, Subhransu Maji, Iasonas Kokkinos, Sammy Mohamed, and Andrea
  Vedaldi.
\newblock Describing textures in the wild.
\newblock In {\em Proceedings of the IEEE Conference on Computer Vision and
  Pattern Recognition}, pages 3606--3613, 2014.

\bibitem{codella2019skin}
Noel Codella, Veronica Rotemberg, Philipp Tschandl, M~Emre Celebi, Stephen
  Dusza, David Gutman, Brian Helba, Aadi Kalloo, Konstantinos Liopyris, Michael
  Marchetti, et~al.
\newblock Skin lesion analysis toward melanoma detection 2018: A challenge
  hosted by the international skin imaging collaboration (isic).
\newblock {\em arXiv preprint arXiv:1902.03368}, 2019.

\bibitem{finn2017model}
Chelsea Finn, Pieter Abbeel, and Sergey Levine.
\newblock Model-agnostic meta-learning for fast adaptation of deep networks.
\newblock In {\em Proceedings of the 34th International Conference on Machine
  Learning-Volume 70}, pages 1126--1135. JMLR. org, 2017.

\bibitem{gilmer2017neural}
Justin Gilmer, Samuel~S Schoenholz, Patrick~F Riley, Oriol Vinyals, and
  George~E Dahl.
\newblock Neural message passing for quantum chemistry.
\newblock In {\em Proceedings of the 34th International Conference on Machine
  Learning-Volume 70}, pages 1263--1272, 2017.

\bibitem{griffin2007caltech}
Gregory Griffin, Alex Holub, and Pietro Perona.
\newblock Caltech-256 object category dataset.
\newblock 2007.

\bibitem{guo2020broader}
Yunhui Guo, Noel~C Codella, Leonid Karlinsky, James~V Codella, John~R Smith,
  Kate Saenko, Tajana Rosing, and Rogerio Feris.
\newblock A broader study of cross-domain few-shot learning.
\newblock ECCV, 2020.

\bibitem{helber2019eurosat}
Patrick Helber, Benjamin Bischke, Andreas Dengel, and Damian Borth.
\newblock Eurosat: A novel dataset and deep learning benchmark for land use and
  land cover classification.
\newblock {\em IEEE Journal of Selected Topics in Applied Earth Observations
  and Remote Sensing}, 12(7):2217--2226, 2019.

\bibitem{kearnes2016molecular}
Steven Kearnes, Kevin McCloskey, Marc Berndl, Vijay Pande, and Patrick Riley.
\newblock Molecular graph convolutions: moving beyond fingerprints.
\newblock {\em Journal of computer-aided molecular design}, 30(8):595--608,
  2016.

\bibitem{krizhevsky2009learning}
Alex Krizhevsky, Geoffrey Hinton, et~al.
\newblock Learning multiple layers of features from tiny images.
\newblock 2009.

\bibitem{mohanty2016using}
Sharada~P Mohanty, David~P Hughes, and Marcel Salath{\'e}.
\newblock Using deep learning for image-based plant disease detection.
\newblock {\em Frontiers in plant science}, 7:1419, 2016.

\bibitem{nakamura2019revisiting}
Akihiro Nakamura and Tatsuya Harada.
\newblock Revisiting fine-tuning for few-shot learning.
\newblock {\em arXiv preprint arXiv:1910.00216}, 2019.

\bibitem{nichol2018first}
Alex Nichol, Joshua Achiam, and John Schulman.
\newblock On first-order meta-learning algorithms.
\newblock {\em arXiv preprint arXiv:1803.02999}, 2018.

\bibitem{pan2009survey}
Sinno~Jialin Pan and Qiang Yang.
\newblock A survey on transfer learning.
\newblock {\em IEEE Transactions on knowledge and data engineering},
  22(10):1345--1359, 2009.

\bibitem{satorras2018few}
Victor~Garcia Satorras and Joan~Bruna Estrach.
\newblock Few-shot learning with graph neural networks.
\newblock In {\em International Conference on Learning Representations}, 2018.

\bibitem{snell2017prototypical}
Jake Snell, Kevin Swersky, and Richard Zemel.
\newblock Prototypical networks for few-shot learning.
\newblock In {\em Advances in neural information processing systems}, pages
  4077--4087, 2017.

\bibitem{triantafillou2017few}
Eleni Triantafillou, Richard Zemel, and Raquel Urtasun.
\newblock Few-shot learning through an information retrieval lens.
\newblock In {\em Proceedings of the 31st International Conference on Neural
  Information Processing Systems}, pages 2252--2262, 2017.

\bibitem{tschandl2018ham10000}
Philipp Tschandl, Cliff Rosendahl, and Harald Kittler.
\newblock The ham10000 dataset, a large collection of multi-source
  dermatoscopic images of common pigmented skin lesions.
\newblock {\em Scientific data}, 5:180161, 2018.

\bibitem{wang2017chestx}
Xiaosong Wang, Yifan Peng, Le Lu, Zhiyong Lu, Mohammadhadi Bagheri, and
  Ronald~M Summers.
\newblock Chestx-ray8: Hospital-scale chest x-ray database and benchmarks on
  weakly-supervised classification and localization of common thorax diseases.
\newblock In {\em Proceedings of the IEEE conference on computer vision and
  pattern recognition}, pages 2097--2106, 2017.

\bibitem{zhang2019variational}
Jian Zhang, Chenglong Zhao, Bingbing Ni, Minghao Xu, and Xiaokang Yang.
\newblock Variational few-shot learning.
\newblock In {\em Proceedings of the IEEE/CVF International Conference on
  Computer Vision}, pages 1685--1694, 2019.

\bibitem{zhou2017places}
Bolei Zhou, Agata Lapedriza, Aditya Khosla, Aude Oliva, and Antonio Torralba.
\newblock Places: A 10 million image database for scene recognition.
\newblock {\em IEEE transactions on pattern analysis and machine intelligence},
  40(6):1452--1464, 2017.

\end{thebibliography}
}

\end{document}